%% file: main.tex
\documentclass[runningheads]{llncs}

\usepackage{enumerate}
\usepackage{amsmath}
\usepackage{amsfonts}
\usepackage{subfigure}
\usepackage{graphicx}
\usepackage[hidelinks]{hyperref}
\usepackage{wrapfig}
\DeclareMathOperator{\score}{ObAlEx}
\DeclareMathOperator{\avgscore}{AvgObAlEx}

\newcommand{\orcid}[1]{\href{https://orcid.org/#1}{\includegraphics[width=10pt]{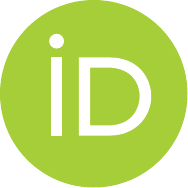}}}

\begin{document}

\title{Right for the Right Reasons: Making Image Classification Intuitively Explainable}
\titlerunning{Making Image Classification Intuitively Explainable}

\author{Anna Nguyen\textsuperscript{\orcid{0000-0001-9004-2092}}\and Adrian Oberf{\"o}ll\and Michael F{\"a}rber\textsuperscript{\orcid{0000-0001-5458-8645}}}

\institute{Karlsruhe Institute of Technology (KIT), Karlsruhe, Germany\\
\email{anna.nguyen@kit.edu}\\
\email{adrian.oberfoell@student.kit.edu}\\
\email{michael.faerber@kit.edu}}

\authorrunning{A. Nguyen et al.}

\maketitle

\begin{abstract}

The effectiveness of Convolutional Neural Networks (CNNs) in classifying image data has been thoroughly demonstrated.
In order to explain the classification to humans, methods for visualizing classification evidence have been developed in recent years.
These explanations reveal that sometimes images are classified correctly, but for the wrong reasons, i.e., based on incidental evidence.
Of course, it is desirable that images are classified correctly for the right reasons, i.e., based on the actual evidence.
To this end, we propose a new \emph{explanation quality metric} to measure \emph{ob}ject \emph{al}igned \emph{ex}planation in image classification which we refer to as the {\itshape ObAlEx} metric.
Using object detection approaches, explanation approaches, and ObAlEx, we quantify the focus of CNNs on the actual evidence.
Moreover, we show that additional training of the CNNs can improve the focus of CNNs without decreasing their accuracy.
\end{abstract}

\input{sections/section1}

\input{sections/section3}

\input{sections/section4}

\input{sections/section5}

\bibliographystyle{splncs04}
\bibliography{References}
\end{document}

%% file: sections/section1.tex
\section{Introduction}

Convolutional Neural Networks (CNNs) have been demonstrated to be very effective in image classification tasks, achieving high accuracy.
However, methods to explain classifications performed by CNNs have shown that sometimes image data has been classified for incidental evidences, undermining the trust between humans and machines~\cite{RSG16}.
Previous attempts to fix this problem have included a human-in-the-loop approach~\cite{SST+20}, a pre-processing step for removing features of the input that are deemed irrelevant for the classification task at hand (such as images' backgrounds)~\cite{JiaLC18}, or the introduction of a new loss function that incorporates an explanation approach during training~\cite{RHD17}.
Although the latter work constrains the explanation of the model in the loss function penalizing the input gradients, it uses explanations only based on input gradients which is not ideal for all use cases, especially in image classification, where individual pixels are difficult to interpret.
Overall, we believe that there is a lack of a metric which quantifies if an intuitive explanation can be gained. 

In this paper, we propose an \emph{ob}ject \emph{al}igned \emph{ex}planation quality metric, called \emph{ObAlEx}.
ObAlEx quantifies to which degree the object mask of an image is consistent with the obtained evidence of explanation methods and thus, imitates human behavior to classify images according to the objects contained.
The proposed metric is independent of the used explanation method (e.g., occlusion~\cite{ZF13}, LIME~\cite{RSG16}, or Grad-cam~\cite{SDV+16}) and object detection method and can therefore be applied together with arbitrary explanation methods and object detection methods.
Our approach to identify the focus on the relevant input regions requires neither human interaction nor pre-processing. 
Based on extensive experiments, we demonstrate the effectiveness of the proposed metric while training CNNs, ensuring both high accuracy and a focus on the relevant input regions.

Our main contributions are as follows:
\begin{enumerate}
 \item We propose an object aligned explanation metric, ObAlEx, to quantify explanations of image classification models intuitively. Our metric is applicable to different explanation methods and neither requires human interaction nor interference in the model's architecture.
 \item In extensive experiments,\footnote{\label{code}We provide the source code online at \url{https://github.com/annugyen/ObAlEx}} 
 we show that our metric can be used for making CNN models for image classification more intuitive while keeping the accuracy.
\end{enumerate}
In the following section, we outline our metric.
We then present our extensive experiments.
Finally, we close with some concluding remarks.

%% file: sections/section3.tex
\section{ObAlEx Metric}
\label{sec:3}

\begin{figure}[tb]
    \centering
    \includegraphics[width = 0.85\linewidth]{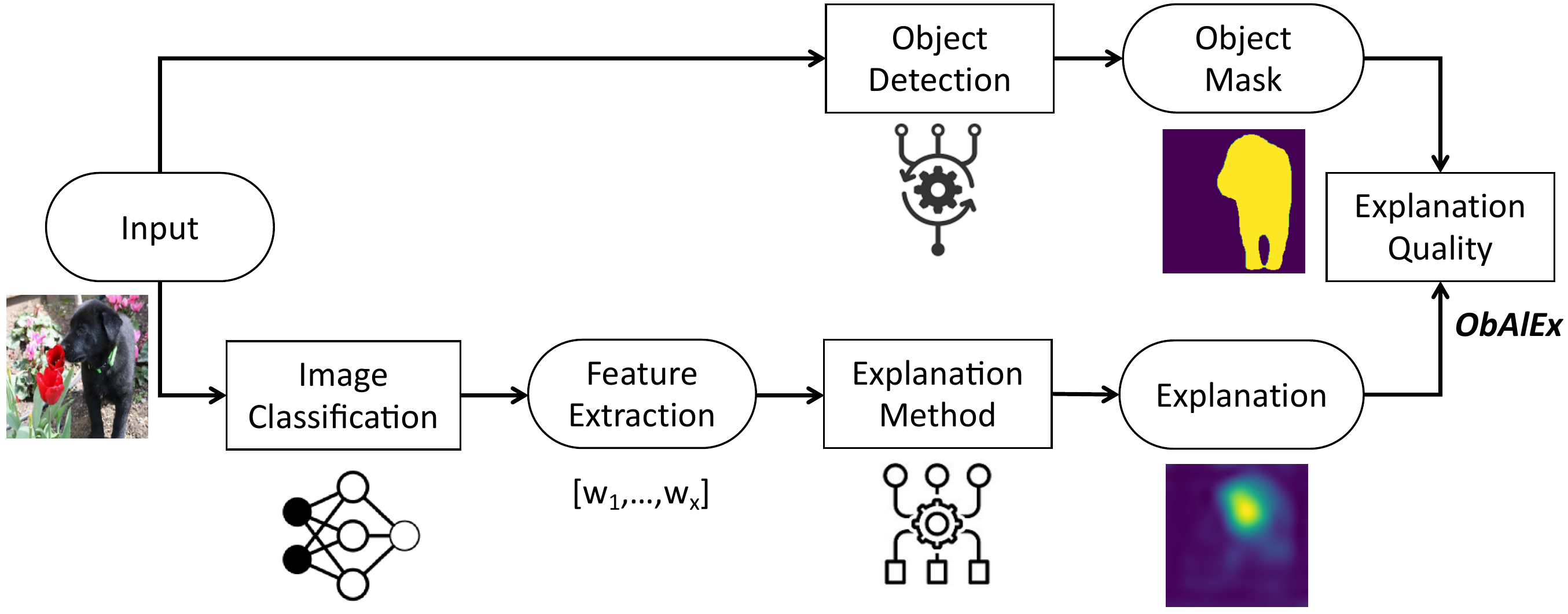}
    \caption{The pipeline of our metric.}
    \label{fig:approaches}
\end{figure}

The metric ObAlEx is designed as a relative metric which depends on the explanation method and the classifier used. Based on the change of the explanation quality during training, it can be evaluated if a certain training strategy leads to an improvement or deterioration of the model's intuitive explanation. By explanation quality, we define the degree of alignment between object to be classified and explanation of the classification model.

The pipeline to calculate ObAlEx is outlined in Fig.~\ref{fig:approaches}. Given an input image on which an object should be detected, we first apply an object detection method (e.g., Mask R-CNN) to obtain the image regions of the object itself (i.e., object mask). We define regions of the explanation that lie outside of the object mask as indicative of a classification for the wrong reasons, and conversely, that regions of the explanation that lie inside of the object mask as indicative of a classification for the right reasons. 
The mask of objects on images can be obtained with a high accuracy nowadays (see Sec.~\ref{sec:4}). 

Simultaneously, an image classifier (e.g., pretrained VGG16) is applied to obtain labels of recognized objects (e.g., "dog"). An explanation method (e.g., Grad-Cam) then outputs the image regions which are most influential given the extracted features from the CNN and the input image.

Both the object mask and the explanation output is then used to compute the metric ObAlEx and thus, to improve the explanation quality. 
Since existing explanation methods support different highlighting levels, our score is constructed in such a way that the score is the higher the more of the highlighted explanation aligns with the object mask. 
In the following, we describe the computation of the explanation quality formally. 

Given a data set $D$ with correctly classified images and an image $d\in D$ with pixels $p^d_{ij}$, width $w^d$, and height $h^d$, let $A^d$ denote the matrix whose values $a^d_{ij}$ equals the activation of the pixels of the object mask, where $i\in\{1,\dots, h^d\}$, $j\in\{1,\dots, w^d\}$, $h^d,w^d\in\mathbb{N}$.
We regard $A^d$ as a fuzzy set, i.e. whose values have degrees of membership depicted as $a^d_{ij}$. We define $a^d_{ij} \in \mathbb{R}$ with $0 \le a^d_{ij} \le 1$. In our experiments, we set $a^d_{ij} = 1$ if the pixel $p^d_{ij}$ of the input image belongs to the object mask and $a^d_{ij} = 0$, otherwise.
Similarly, let $B^d$ be the matrix whose values $b^d_{ij}$ equals the activation of the pixels of the explanation. We additionally normalize the values $b^d_{ij}$ between zero and one, i.e. $0 \le b^d_{ij} \le 1$ where $b^d_{ij} = 1$ if the pixel $p^d_{ij}$ of the input image belongs to the highest activation and $b^d_{ij} = 0$ otherwise. Our metric ObAlEx is, then, defined as follows:
\begin{equation}
    \score (A^d,B^d) = \frac{\sum_{i,j}a^d_{ij}b^d_{ij}}{\sum_{i,j}b^d_{ij}} \in[0,1]
\end{equation}


To get the explanation quality of an image classifier, ObAlEx can be applied on all images in a data set $D$. We then calculate the average of all values of the explanation quality of each picture for an image collection. In doing so, we weight all images equally. The explanation quality of the classifier is defined as 
\begin{equation} 
\label{eq:avg_score}
    \avgscore(D) = \frac{1}{n} \sum_{d=1}^{n} \score(A^d,B^d) \in [0, 1],
\end{equation}
where $n\in\mathbb{N}$ is the number of images in data set $D$. 
$\avgscore$ only considers the scores of images classified correctly by the model, otherwise the metric would get skewed. Therefore, images which are classified wrong are excluded.

%% file: sections/section4.tex
\newpage
\section{Evaluation}
\label{sec:4}

\subsection{Evaluation Setting}
\label{sec:4_2}

To evaluate ObAlEx, we apply pre-trained CNN models. 
We focus on three state-of-the-art image classification models:
\emph{VGG16} \cite{SZ15}, \emph{ResNet50} \cite{HZR+15}, and \emph{MobileNet} \cite{HZC+17}. 
The models are pre-trained on the ILSVRC2012 data set~\cite{ILSVRC15} which is also known as ImageNet. We adapt each model's upper output dense layers to the specific data set (i.e., number of categories in the used image classification data sets \emph{Dogs vs. Cats} and \emph{Caltech 101}, respectively).
To show the universal applicability of ObAlEx, we use different well-known explanation methods such as occlusion~\cite{ZF13}, LIME~\cite{RSG16}, Grad-Cam~\cite{SDV+16}, and Grad-Cam++~\cite{CSH+17}.
In our experiments, the $\avgscore$ settled around a fixed value after 50 images. For that reason and due to high computing power costs in case of LIME, we calculate the $\avgscore$ for 50 images per epoch in the following experiments.
Our experiments are executed on a server with 12~GB of GPU RAM. We use TensorFlow and the Keras deep learning library for implementation. We use the following data sets in our evaluation:

\paragraph{Dogs vs. Cats}
data set\footnote{\url{https://www.kaggle.com/c/dogs-vs-cats}, last accessed: 2020-10-28} contains 3,000 dog and cat images, 1,500 per class. 
We use Mask R-CNN~\cite{HGD+17} to create the object masks.
The quality of the object masks is important for the validity of the proposed metric ObAlEx. Therefore, we manually evaluated the computed object masks for 200 randomly chosen images regarding the overlap of the whole object. 
The accuracy was 91\%. 
Thus, we argue that the pre-trained Mask R-CNN performs well for our purpose. 

Given the data set size, we used 70\% of the images for training and 30\% for testing.
We first adjust the output layer of all CNN models to the two categories (dog and cat) and train them for 10 epochs on the Dogs vs. Cats data set (where all layers except output layer are frozen).
After that, we freeze different combinations of layers for further training. 
In the original papers of the above mentioned models, the convolutional layers are divided into five blocks. For simplification and comparability, we use this convention for our strategies. We also summarize the last dense layers to one block. Thus, we always set whole blocks of layers to either be trainable or non-trainable. We train every strategy for another 10 epochs.
We investigate the following strategies: (a) train the last dense layers which we denote as dense block, (b) train the last two convolutional blocks (i.e. the fourth and fifth), (c) train the first three convolutional blocks, and (d) train all layers, i.e. all convolutional and dense blocks.

\begin{figure*}
    \centering
    \includegraphics[width = 0.85\linewidth]{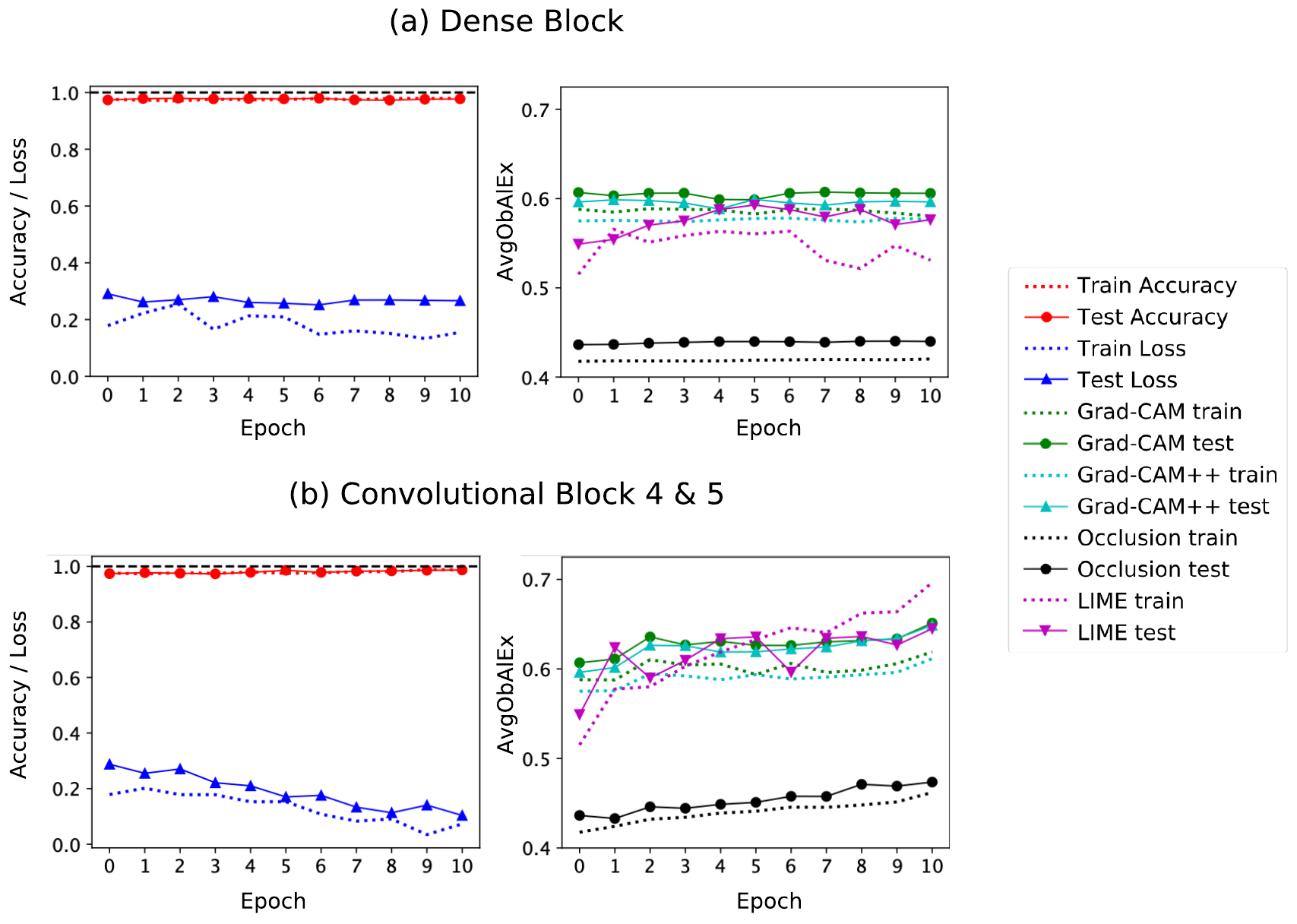}
    \vspace*{-0.2cm}
    \caption{VGG16 Results. Transfer learning strategies with VGG16 with explanation methods Occlusion, LIME and Grad-Cam/Grad-Cam++.}
    \label{fig:vgg_results}
\end{figure*}

\paragraph{Caltech 101}
data set~\cite{FFP07} has 101 object categories.
We create a uniform distributed data set by drawing random sampling from the categories resulting in a total of 6,060 images with 60 images per class. We use a test split of 0.25. 
This data set is provided with hand-labeled object masks for all images. Thus, we use those labeled object masks.
We perform another experiment inspired by~\cite{RHD17,SST+20}. To actively force the model to be more intuitive and thus, to provide a more interpretable explanation, we followed a na\"ive approach by using artificial images. We edit the images in a way that they contain the object to classify and masked out the background with random pixels. This should force the model to focus more on the object and increase the explanation quality.

\subsection{Evaluation Results}

\paragraph{Dogs vs. Cats.} 
Fig.~\ref{fig:vgg_results} shows the results for VGG16 with training strategies (a) and (b). We can see that the performance of the model measured with accuracy did not change within 10 epochs (see Fig.~\ref{fig:vgg_results} (a)/(b) left graph). However, we observed a change in $\avgscore$ (see Fig.~\ref{fig:vgg_results} (a)/(b) right graph). 
The explanation quality after 10 epochs computed with any explanation method for strategy (b) is significantly higher than the explanation quality for strategy (a). This fits to the common knowledge that complex structures in the input images are learned in the later convolutional blocks and are, therefore, more decisive for the classification. 
Moreover, Fig.~\ref{fig:vgg_results} (b) shows with increasing number of epochs a decrease in the loss, while the $\avgscore$ increases simultaneously. This indicates the effectiveness of the model for right predictions based on the right reasons.
The results of strategy (d) and (b) and the results of strategy (c) and (a) are similar to each other respectively, which emphasizes the common knowledge. Without using the proposed metric ObAlEx this improvement would not be evident since the accuracy of all models stays the same during training. 

\begin{wrapfigure}{l}{4.7cm}
    \centering
    \vspace{-0.2cm}
    \begin{subfigure}[]{\includegraphics[width=0.45\linewidth]{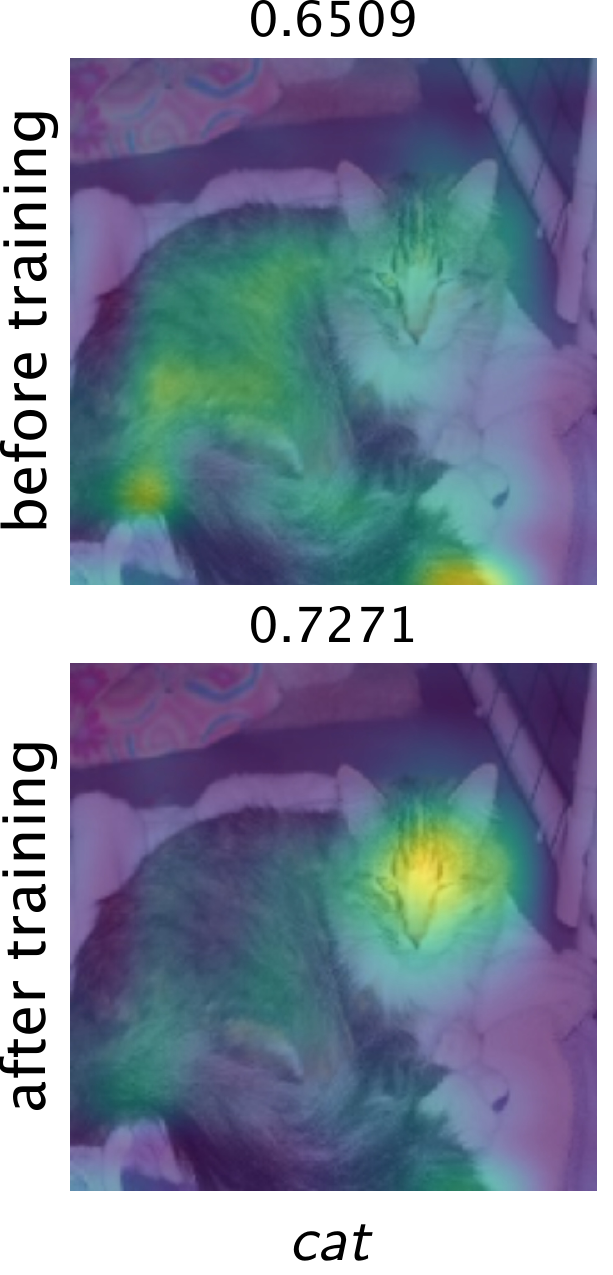}}
    \label{sfig:dvc}
    \end{subfigure}
    \begin{subfigure}[]{\includegraphics[width=0.45\linewidth]{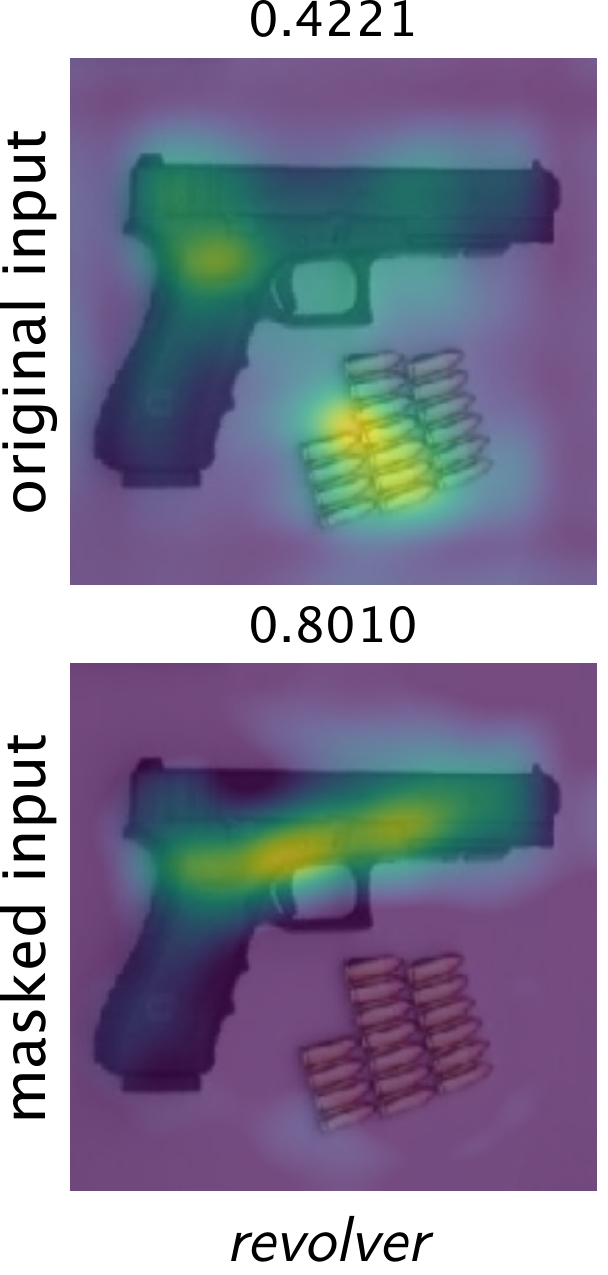}}
    \label{sfig:caltech}
    \end{subfigure}
   \vspace*{-0.3cm}
\caption{Examples from (i)~Dogs vs. Cats and (ii)~Caltech 101 with quality scores shown above.}
\label{fig:exp_vgg16}
\vspace{-1cm}
\end{wrapfigure}

In Fig.~\ref{fig:exp_vgg16}~(i) , we provide an example of the explanation visualized with Grad-Cam with strategy (b) on VGG16. We can see that the explanation quality increases after training and that the visualized explanation has a stronger focus on the object. 
With only 10 epochs of additional training, we were able to improve the model in a way that it utilizes more important features such as the face of the animal.
Without ObAlEx, it would be obvious to not train the model any further due to the non-changing accuracy. 
We observed similar results on the experiments with ResNet50 and MobileNet, and also on the Caltech 101 data set but omit them due to page limitations.\textsuperscript{\ref{code}}

\begin{wrapfigure}{l}{7cm}
    \centering
    \vspace{-0.7cm}
    \includegraphics[width = \linewidth]{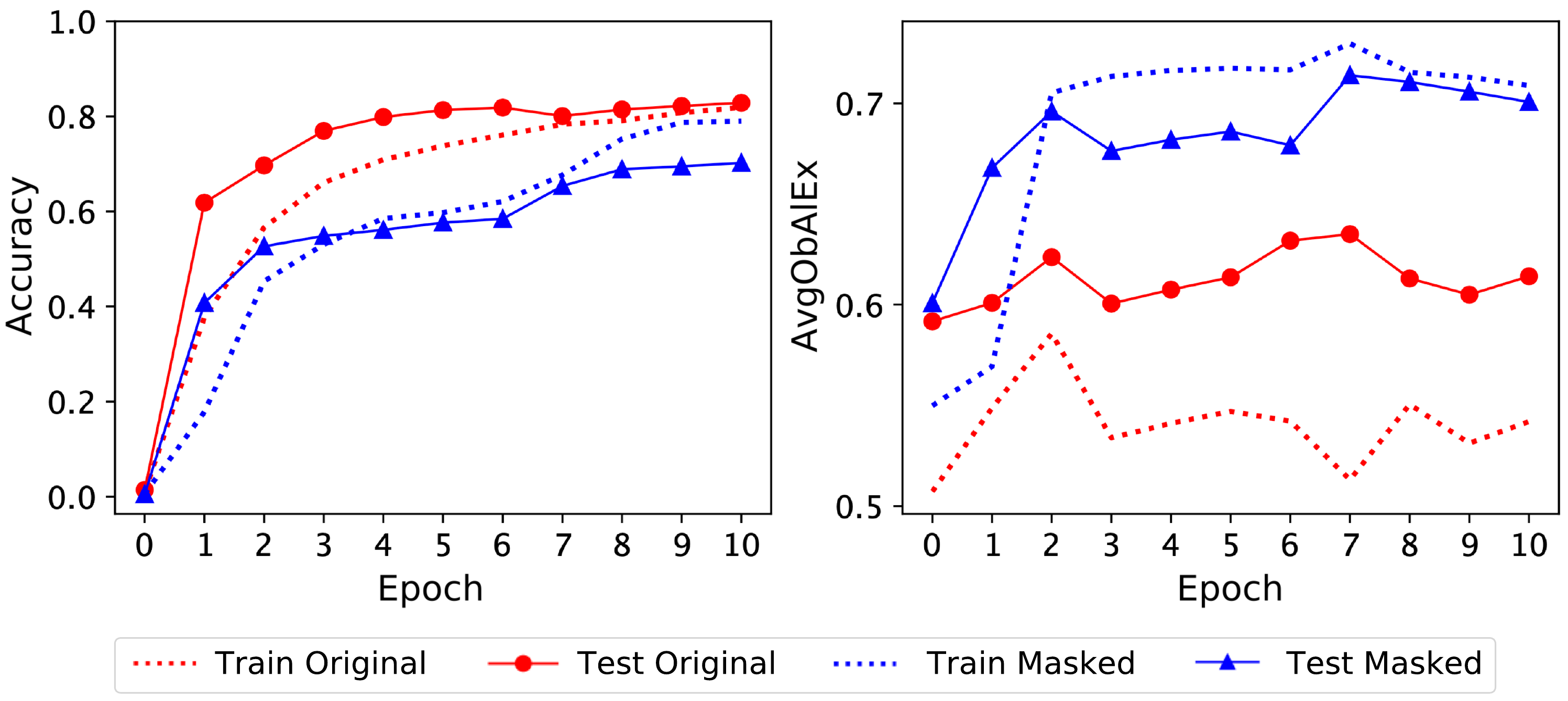}
    \vspace*{-0.5cm}
    \caption{Training on Caltech 101.}
    \label{fig:masked_caltech}
    \vspace{-0.7cm}
\end{wrapfigure}

\paragraph{Caltech 101}
Fig.~\ref{fig:masked_caltech} shows the results for 10 epochs of training VGG16 on Caltech 101 with the original and masked images as input. As we can observe in the left 
graph, training with the original images results in a higher accuracy than training with the masked images. However, the $\avgscore$ (computed with Grad-Cam as explainer, see graph on the right) of the model trained with masked input images is significantly higher than the $\avgscore$ of the model trained with the original input images. 
This indicates that more background information was used in the classification. Thus, evaluating image classifiers beyond accuracy can be valuable to real-world cases where specific background information is unavailable.

Fig.~\ref{fig:exp_vgg16} (ii) shows an example image with Grad-Cam on VGG16. Despite high accuracy, we can see that the explanation for the image with masked out background (image at the bottom) is more intuitive and more focused on the actual object than the original input image.

%% file: sections/section5.tex
\section{Conclusion}
\label{sec:5}

In this paper, we focused on evaluating CNN image classifiers with different explanation approaches. We introduced a novel explanation quality score metric to support the training process besides accuracy and loss function. We have shown in our experiments that our metric ObAlEx can be used to indicate cases where a model makes its predictions based on wrong reasons. Overall, ObAlEx facilitates more generalized models which can increase the user's trust in the model by object aligned explanations.